\documentclass{article}
\usepackage{spconf,graphicx}
\usepackage[caption=false]{subfig}
\usepackage{amsmath,amssymb,amsfonts}
\usepackage{comment}
\usepackage{xcolor}

\title{Kernelized dense layers for facial expression recognition}
\newcommand\Mark[1]{\textsuperscript#1}

\name{M.Amine Mahmoudi\Mark{1}\Mark{,}\Mark{3}, Aladine Chetouani\Mark{2}, Fatma Boufera\Mark{1}, Hedi Tabia\Mark{3} }
\address{\Mark{1}Mustapha Stambouli University of Mascara, Algeria \\
\Mark{2}PRISME laboratory, University of Orleans, Orleans, France\\
\Mark{3}IBISC laboratory, University of Paris-Saclay, Paris, France\\
mohamed.mahmoudi@univ-mascara.dz}

\begin{document}
%
\maketitle
\begin{abstract}

Fully connected layer is an essential component of Convolutional Neural Networks (CNNs), which demonstrates its efficiency in computer vision tasks. The CNN process usually starts with convolution and pooling layers that first break down the input images into features, and then analyze them independently. The result of this process feeds into a fully connected neural network structure which drives the final classification decision. In this paper, we propose a Kernelized Dense Layer (KDL) which captures higher order feature interactions instead of conventional linear relations. We apply this method to Facial Expression Recognition (FER) and evaluate its performance on RAF, FER2013 and ExpW datasets. The experimental results demonstrate the benefits of such layer and show that our model achieves competitive results with respect to the state-of-the-art approaches.
\end{abstract}
\section{Introduction}
\label{sec:intro}

Facial Expression Recognition (FER) research aims at classifying the human emotions given from facial images as one of seven basic emotions: happiness, sadness, fear, disgust, anger, surprise and neutral. FER finds applications in different fields including security, intelligent human-computer interaction, and clinical medicine. Recently, many FER works~\cite{kim2016fusing,liu2017adaptive,li2017reliable,zhang2018facial,zeng2018facial,yang2018facial,acharya2018covariance} based on CNNs have been proposed in the literature. Although these works mainly differ by the model architectures and the used databases, convolution and Fully Connected (FC) layers are often employed.

The classic neural network architecture was found to be inefficient for computer vision tasks, since images represent a large input for a neural network (they can have hundreds or thousands of pixels and up to 3 color channels) with a huge number of connections and network parameters. CNNs leverage the fact that an image is composed of smaller details, or features, and creates a mechanism for analyzing each feature in isolation, which informs a decision about the image as a whole. As part of the convolutional network, FC layer uses the output from the the convolution/pooling process and learns a classification decision. It is an essential component of CNNs, which demonstrates its utility in several computer vision tasks. The input values flow into the first FC layer and they are multiplied by weights. The latter usually go through an activation function (typically ReLu), just like in a classic artificial neural network. They then pass forward to the output layer, in which every neuron represents a classification label. The FC part of the CNN goes through its own back-propagation process to determine the most accurate weights where each neuron receives weights that prioritize the most appropriate label.

To improve the performance of CNNs, several methods using higher order kernel function than the ordinary linear kernel have been proposed in the literature. In~\cite{cui2017kernel}, a novel pooling method in the form of Taylor series kernel has been proposed. This method captures high order and non-linear feature interactions via compact explicit feature mapping. The approximated representation is fully differentiable, thus the kernel composition can be learned together with a CNN in an end-to-end manner. It acts as a basis expansion layers, increasing thereby the discrimination power of the FC layers. These methods have attracted increasing attentions, achieving better performance than classical first-order networks in a variety of computer vision tasks. Wang et al.~\cite{wang2019kervolutional} focused more one the convolution part and they proposed to replace the convolution layers in a CNN by kernel-based layers, called kervolution layers. The use of these layers increases the model capacity to capture higher order features at the convolutional phase.

In this paper, we build upon these works and introduce a novel FC layer. We leverage kernel functions to build a neuron unit that applies a higher order function on its inputs instead of calculating their weighted sum. The proposed Kernelized Dense Layers (KDL) permits to improve the discrimination power of the full network and it is completely differentiable, allowing an end-to-end learning. The experimental results demonstrate the benefits of such layer in FER task and show that our model achieves competitive results with respect to the state-of-the-art approaches.

The remainder of this paper is organized as follow: Section~\ref{sec:KDLL} introduces the proposed KDL for FER. Section~\ref{sec:Experiments} presents the different conducted experiments and their related results. Section~\ref{sec:Conclusion} concludes the paper.


\section{kernelized Dense Layer}
\label{sec:KDLL}

The proposed kernelized Dense Layer is similar to a classical neuron layer in the way that it applies a dot product between a vector of weights and an input vector, add a bias vector ($b\geq 0$) and eventually applies an activation function. The difference from standard FC layers is that our proposed method applies higher degree kernel function instead of a simple linear dot product, which allows the model to map the input data to a higher space and thus be more discriminative than a classical linear layer. 

\begin{figure}[ht]
\begin{center}
\includegraphics[width=\linewidth]{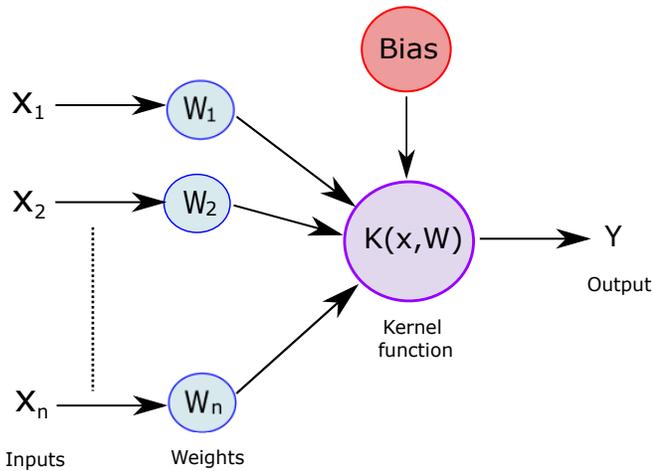}
\end{center}
\caption{The basic unit of our proposed KDL is a kernel neuron. It applies a kernel function on an input vector $x =\{x_1,x_2,\dots,x_n\}$ and a vector of weights $w = \{w_1,w_2,\dots,w_n\}$, adds a bias term and eventually applies an activation function.}
\label{fig:Kernel_Neuron}
\end{figure}

Figure \ref{fig:Kernel_Neuron} shows the processing of an elementary unit (kernel neuron) of our proposed KDL. Formally, the output $Y$ is computed by applying a kernel function $K$ on an input vector $x =\{x_1,x_2,\dots,x_n\}$ and the corresponding vector of weights $W = \{w_1,w_2,\dots,w_n\}$ and, adding the bias vector ($b\geq 0$).

In this work, we employed two different kernel functions which have an Euclidean structure $\mathbb{R}^{d}$. 

\begin{itemize}
    \item \textbf{Linear kernel:}
        \begin{equation}
            \mathcal{K}(x,w)=x^{T}w+b,\quad x,w\in \mathbb {R} ^{d},b\geq 0.
            \label{eq:LinearKernel}
        \end{equation}
\end{itemize}

The linear kernel (Equation~\ref{eq:LinearKernel}) looks at the similarity between the input vector $x$ and the filter weight vector $w$. 

\begin{itemize}
    \item  \textbf{Polynomial kernel:}
        \begin{equation}
            \mathcal{K}(x,w)=(x^{T}w+b)^{n},\quad x,w\in \mathbb{R}^{d},b\geq 0.
            \label{eq:PolynomialKernel}
        \end{equation}
\end{itemize}

Starting form $n>1$ in Equation~\ref{eq:PolynomialKernel}, the polynomial kernel $K$ encodes not only the linear relation between both $x$ and $w$ vectors, but also non-linear relations between them. It corresponds to an inner product in a feature space based on some mapping $\varphi$:
\begin{equation}
    \langle \varphi(x) , \varphi(w)\rangle \approx \mathcal{K}(x,w).
\end{equation}
Note that in the case of polynomial kernel with degree $n>1$, we do not apply an activation function on the neuron output, since non linearity is already added by the high polynomial kernel degree. \\

Other kernels that expand the non-linearity to the infinity can also be used. For instance, the Gaussian kernel defined by $\mathcal{K}(x,w)=e^{-{\frac {\|x-w\|^{2}}{2\sigma ^{2}}}},\quad x,w\in \mathbb {R} ^{n},\sigma >0$, Laplacian kernel defined by $ \mathcal{K}(x,w)=e^{-\alpha \|x-w\|}$, or the Abel kernel defined as $ \mathcal{K}(x,w)=e^{-\alpha |x-w|},\text{ where, }\alpha >0$ in both kernels.

\subsection{Datasets}
\label{sec:datasets}
Our experiments have been conducted on three well-known facial expression datasets: RAF-DB~\cite{li2017reliable}, ExpW~\cite{zhang2018facial} and FER2013~\cite{goodfellow2013challenges}. 

\begin{itemize}
    \item The RAF-DB~\cite{li2017reliable} stands for the Real-world Affective Face DataBase. It  is  a  real-world  dataset  that  contains  29,672  highly diverse  facial  images.
    \item The ExpW~\cite{zhang2018facial} stands for the  EXPression  in-the-Wild  dataset. It contains  91,793 facial  images.
    \item The FER2013 database was first introduced during the ICML 2013 Challenges in Representation Learning \cite{goodfellow2013challenges}. This database contains 28709 training images, 3589 validation images and 3589 test images.
\end{itemize}

\subsection{Training process}

The only pre-processing which we have employed on all experiments is cropping the face region and resizing the resulting images to $100 \times 100$ pixels. We have used Adam optimiser with a learning rate varying from 0.001 to 5e-5. This learning rate is decreased by a factor of 0.63 if the validation accuracy does not increase over ten epochs. To avoid over-fitting we have first augmented the data using a range degree for random rotations of 20, a shear intensity of 0.2, a range for random zoom of 0.2 and randomly flip inputs horizontally. We have also employed earl stopping if validation accuracy does not improve by a factor of 0.01 over 20 epochs. Each KDL of our model is initialized with $He$ normal distribution and a weight decay of 0.0001.

\section{Experiments}
\label{sec:Experiments}

\begin{figure*}[ht]
\begin{center}
\includegraphics[width=\linewidth]{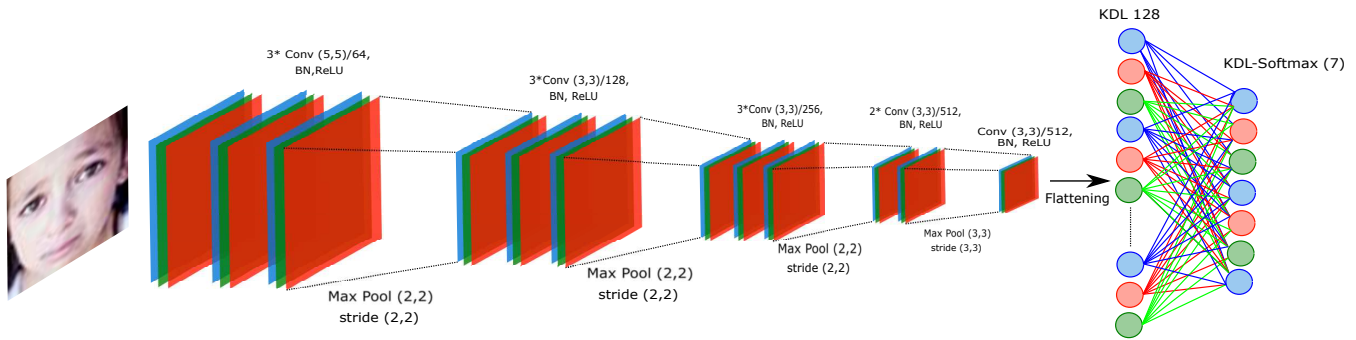}
\end{center}
\caption{Base model architecture: it is composed of five convolutional blocks. Each block consists of a convolution, batch normalization and rectified linear unit activation layers. Each of the five convolutional blocks is followed by a dropout layer. Finally, two KDL are added on top of these convolution blocks with respectively 128 units and ReLU activation and 7 units with softmax activation.}
\label{fig:Base_Model}
\end{figure*}

In order to demonstrate the efficiency of the proposed KDL, we built a simple CNN from scratch rather than using a pre-trained one. As shown in Figure~\ref{fig:Base_Model}, our model architecture is composed of five convolutional blocks. Each block consists of a convolution, batch normalization and rectified linear unit activation layers. The use of batch normalization~\cite{zou2019integration} before the activation brings more stability to parameter initialization and achieves higher learning rate. Each of the five convolutional blocks is followed by a max pooling layer and a dropout layer. Finally, two KDL are added on top of these convolution blocks with respectively 128 and 7 units. On the former, we apply a ReLU activation function in the case of linear kernel only, since non-linearity is already added by the polynomial kernel. Softmax activation function is finally applied on the last KDL.

\subsection{Ablation Study}
\label{sec:AblationStudy}

This section explores the impact of the use of the proposed KDL on the overall accuracy of a CNN model. We evaluated the performance using the same network architecture with different FC techniques. First, we used our model with standard FC layers which gives the same results as kernel neuron layers with a polynomial function with degree (n=1). After that, we replaced these FC layers by our KDL. We studied the behaviour of two different kernel functions, namely; the second-order polynomial kernel and the third-order polynomial kernel. The experiments are conducted with the same training parameters as described above.

\begin{table}[ht]
\caption{Accuracy Rates of the proposed approach }
\begin{center}
\resizebox{\linewidth}{!}{
\begin{tabular}{|c|c|c|c|}
\hline
\textbf{}&\multicolumn{3}{|c|}{\textbf{Dataset}} \\
\cline{2-4} 
\textbf{Models} & \textbf{\textit{FER2013}}& \textbf{\textit{ExpW}}& \textbf{\textit{RAF-DB}} \\
\hline
Base-Model-FC & 70.13\% & 75.91\% & 87.05\% \\
\hline
Base-Model-KDL$^{\mathrm{a}}$ (n=1)& 70.09\% & 76.87\% & 87.03\% \\
\hline
Base-Model-KDL$^{\mathrm{a}}$ (n=2)& 70.85\% & 76.13\% & 87.64\% \\
\hline
Base-Model-KDL (n=3)& \textbf{71.28\%} & \textbf{76.64\%} & \textbf{88.02\%} \\
\hline
\multicolumn{4}{l}{$^{\mathrm{a}}$KDL: Kernelized Dense Layer.}
\end{tabular}
}
\label{tab1}
\end{center}
\end{table}

\begin{figure}[tbp]
	\subfloat[Validation accuracy]{\includegraphics[width=0.45\linewidth]{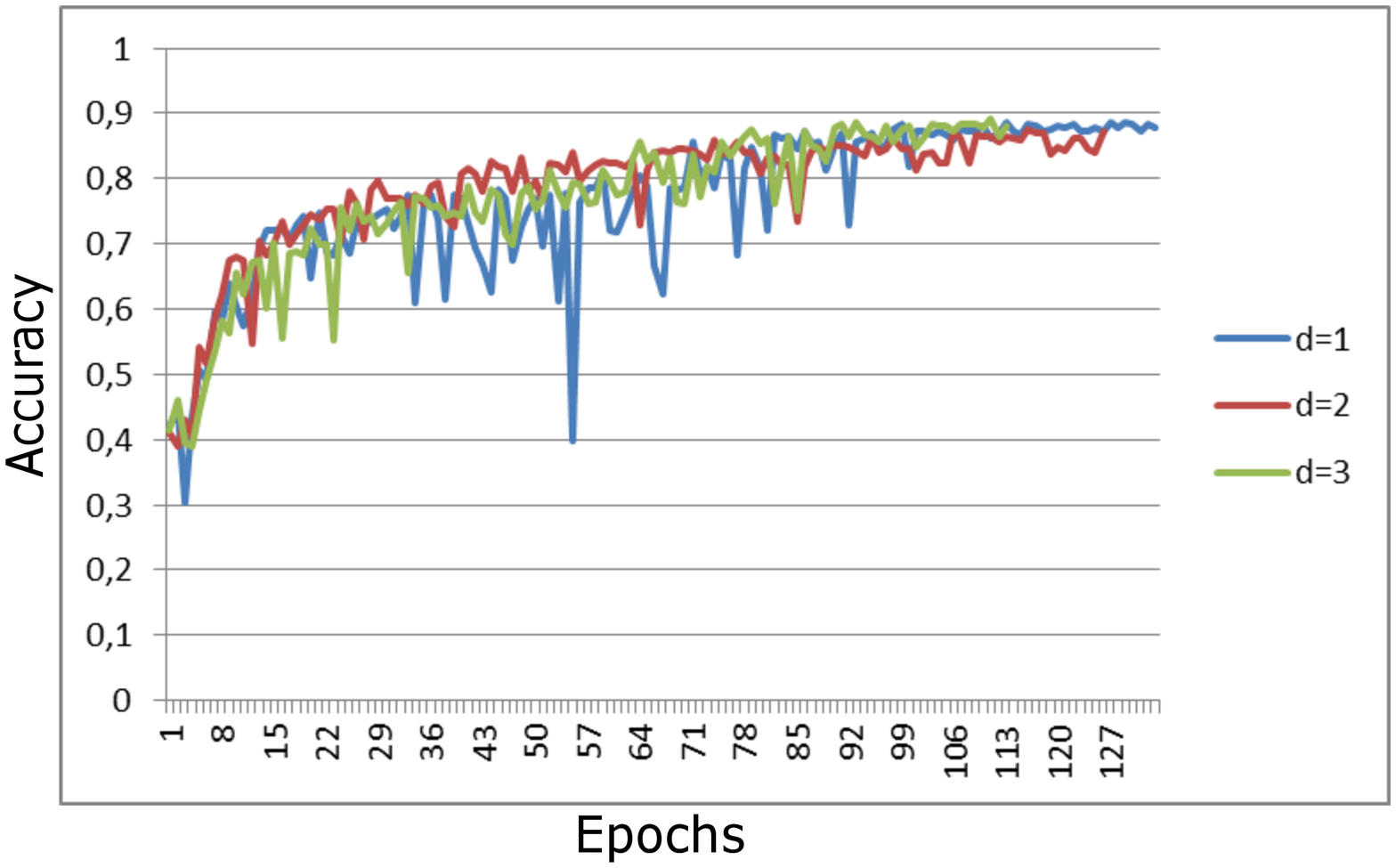}\label{fig:val_acc_RAF_DB}}\
	\subfloat[Validation loss]{\includegraphics[width=0.45\linewidth]{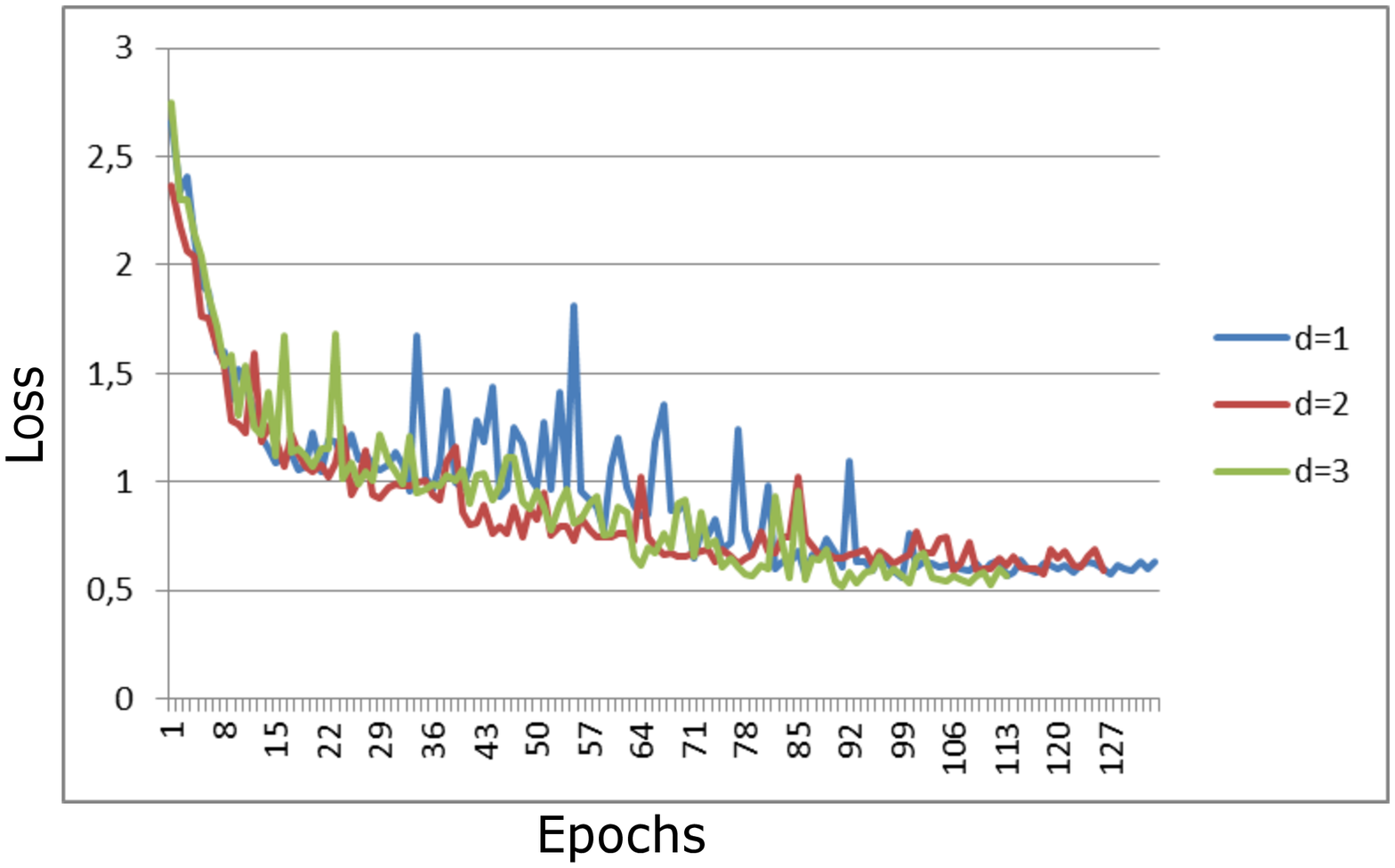}\label{fig:val_loss_RAF_DB}}\\	
	\caption{Validation accuracy and validation loss on RAF-DB with the three kernel configurations.}
	\label{fig:val_acc_loss_RAF_DB}
\end{figure}

Table~\ref{tab1} presents the results of our model using standard FC layers with comparison to the proposed KDL. In the case of the standard FC layers (Base-Model-FC), our base model attains 70.13\%, 75.91\% and 87.05\% of accuracy rate on respectively FER2013, ExpW and RAF datasets, while the use of KDL considerably increases its accuracy. Indeed, one can notice that the accuracy for the second-order polynomial kernel increases for about 0.7\% for FER2013, 0.25\% for ExpW and 0.6\% for RAF-DB. In the same way, using the third-order polynomial kernel increases further the overall accuracy. Compared to standard FC layers, the third-order polynomial KLD enhances the model accuracy for about 1.15\% for FER2013, 0.75\% for ExpW and 1\% for RAF-DB. These results are consistent with previous work \cite{wang2019kervolutional}, where the third-order applied on convolution layers gave the best performance. Although the computational complexity of these kernels is higher compared to the standard layers, they allow to strongly improve the model accuracy. These results demonstrate that the use of KDL, in the case of FER problem, are beneficial for the overall accuracy of the model. These techniques enhance the discriminative power of the model, compared to a standard FC layer.

\begin{figure}[tbp]
	\centering
	\subfloat[Validation accuracy]{\includegraphics[width=0.45\linewidth]{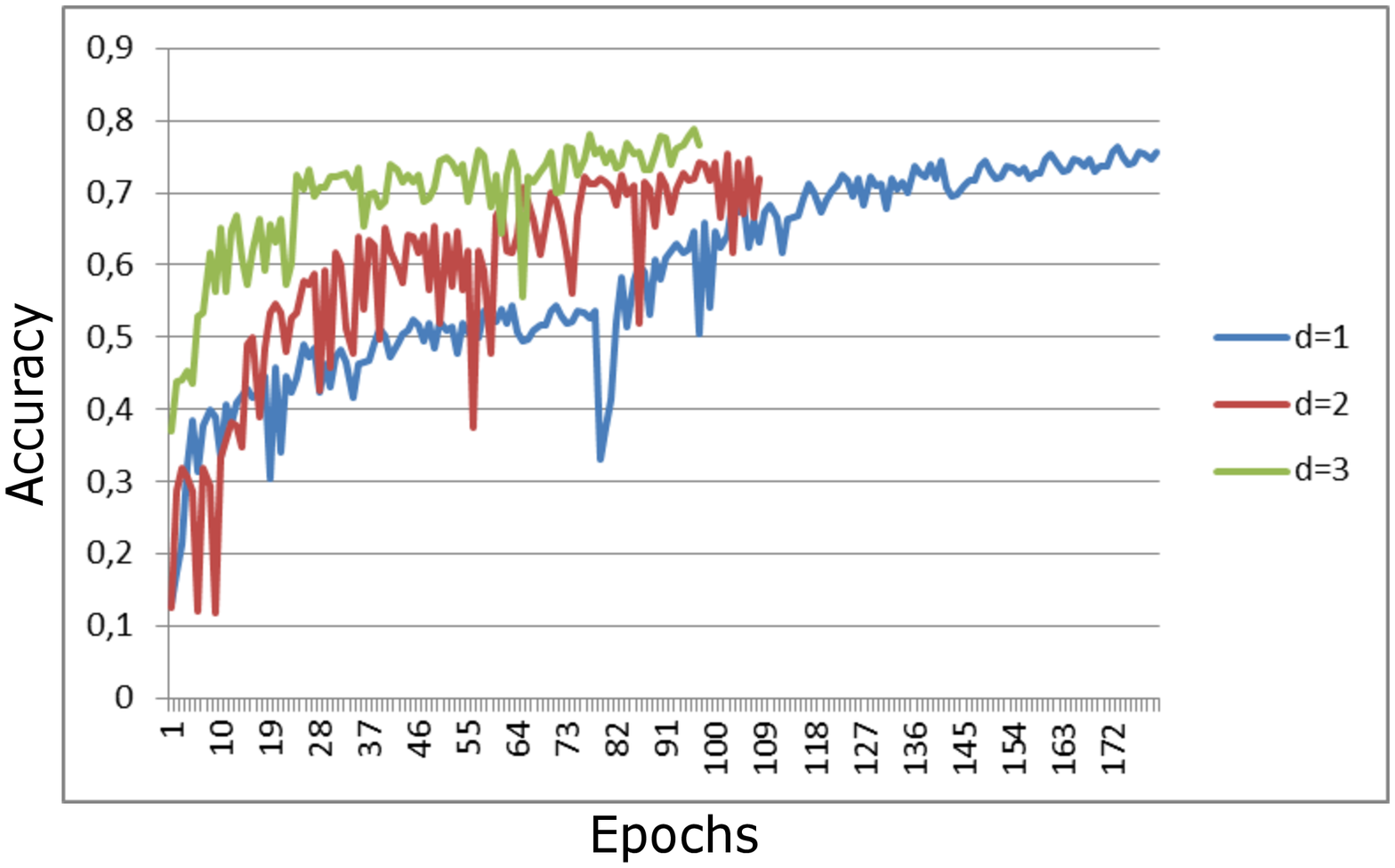}\label{fig:val_acc_ExpW}}\
	\subfloat[Validation loss]{\includegraphics[width=0.45\linewidth]{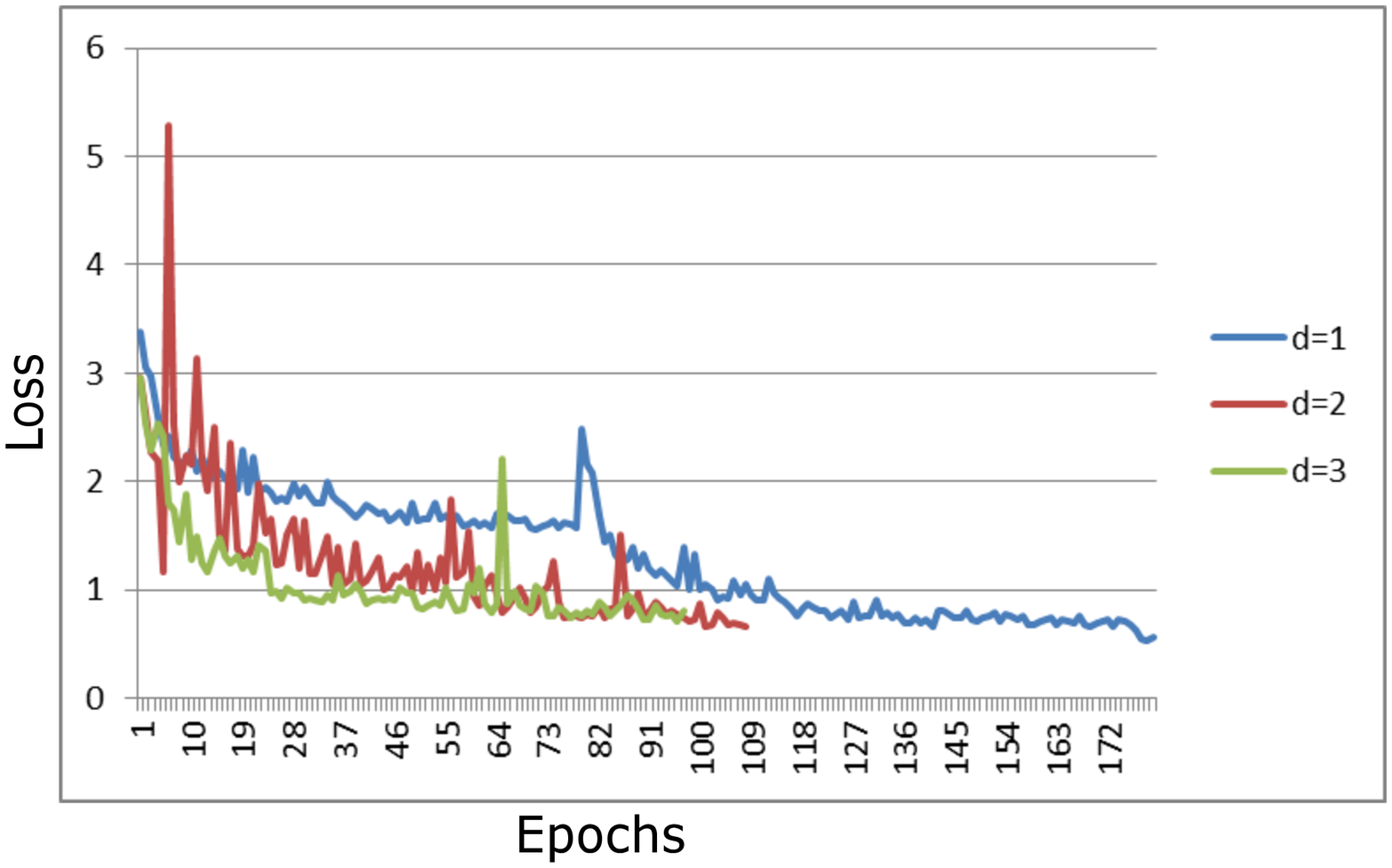}\label{fig:val_loss_ExpW}}\\	
	\caption{Validation accuracy and validation loss on ExpW with the three kernel configurations.}
	\label{fig:val_acc_loss_ExpW}
\end{figure}

\begin{figure}[tbp]
	\centering
	\subfloat[Validation accuracy]{\includegraphics[width=0.45\linewidth]{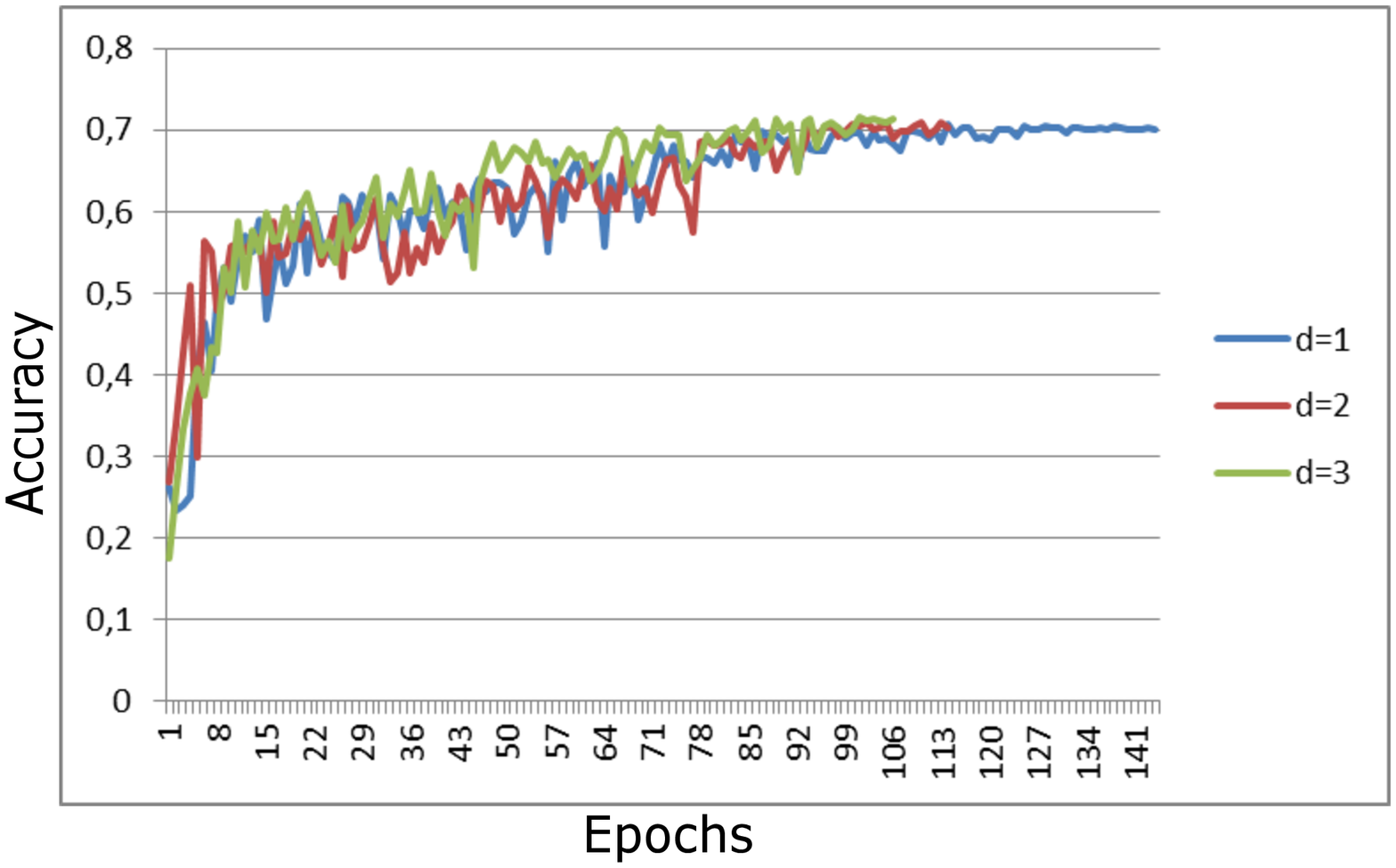}\label{fig:val_acc_FER2013}}\
	\subfloat[Validation loss]{\includegraphics[width=0.45\linewidth]{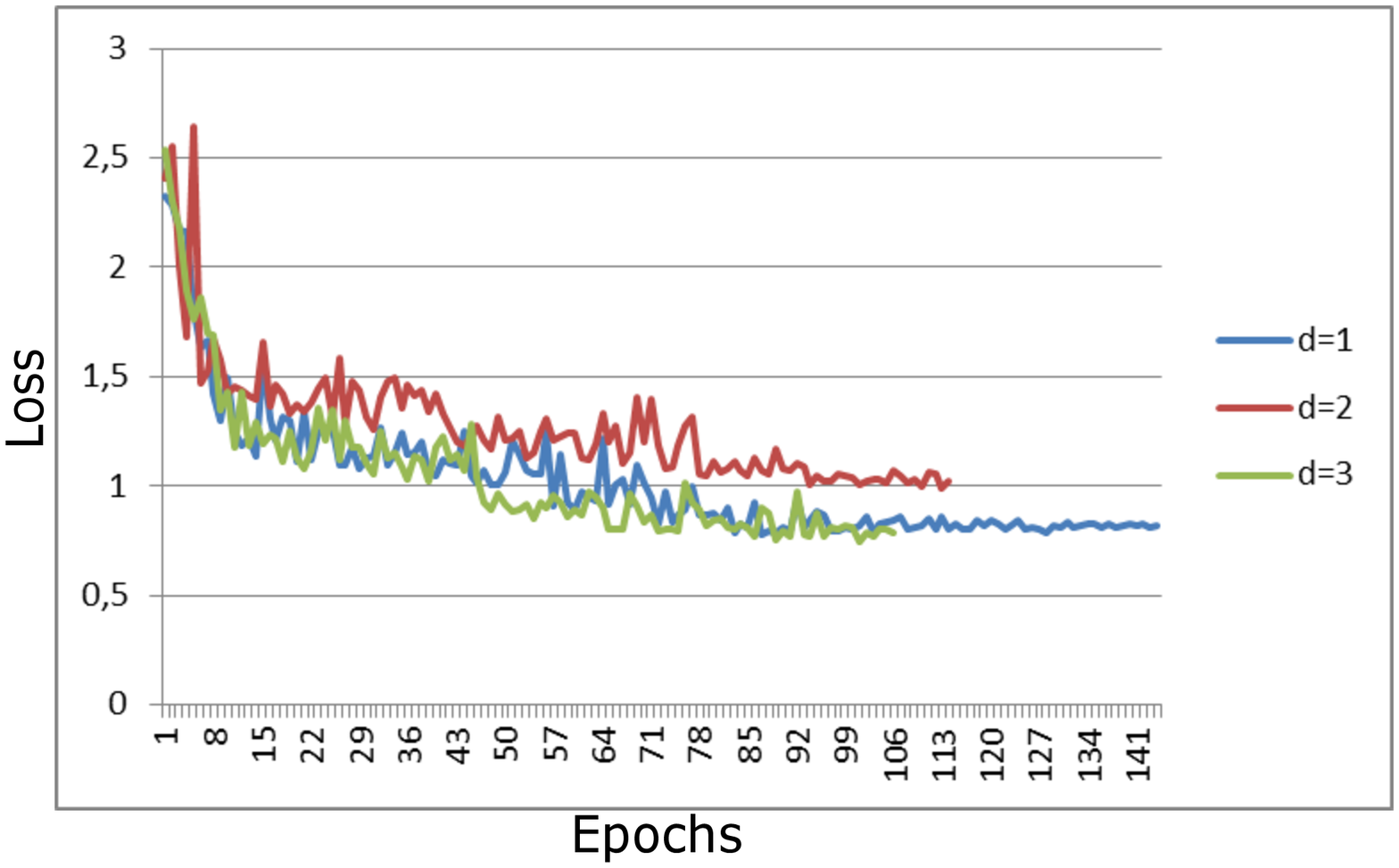}\label{fig:val_loss_FER2013}}\\	
	\caption{Validation accuracy and validation loss on FER2013 with the three kernel configurations.}
	\label{fig:val_acc_loss_FER2013}
\end{figure}

Another beneficial aspect of using KDL is the speed of convergence. As shown in Figures \ref{fig:val_acc_loss_RAF_DB}, \ref{fig:val_acc_loss_ExpW} and \ref{fig:val_acc_loss_FER2013}, the higher degree is the kernel function, the fast it converge. Due to the use of early stopping in our training process, the learning process is interrupted as soon as the model begins to overfit. As can be seen, the higher is the kernel function degree, the sooner it stops training (the blue, red and green curves correspond, respectively, to n=1, n=2 and n=3). High degree kernel functions are known to be prone to overfitting, which in our case limits the number of units and layers of our proposed KDL.

\begin{table}[ht]
\caption{Accuracy rate of the proposed approach and state-of-the-art approach}
\begin{center}
\resizebox{\linewidth}{!}{
\begin{tabular}{|c|c|c|c|}
\hline
\textbf{}&\multicolumn{3}{|c|}{\textbf{Dataset}} \\
\cline{2-4} 
\textbf{Models} & \textbf{\textit{FER2013}}& \textbf{\textit{ExpW}}& \textbf{\textit{RAF-DB}} \\
\hline
Base-Model-FC & 70.13\% & 75.91\% & 87.05\% \\
\hline
Base-Model-KDL$^{\mathrm{a}}$ (n=1)& 70.09\% & 76.87\% & 87.03\% \\
\hline
Base-Model-KDL$^{\mathrm{a}}$ (n=2)& 70.85\% & 76.13\% & 87.64\% \\
\hline
Base-Model-KDL (n=3)& 71.28\% & \textbf{76.64}\% & \textbf{88.02}\% \\
\hline
Tang et al.~\cite{tang2013deep}  &   71.16\% & --   &   --  \\
\hline
Guo et al.~\cite{guo2016deep}& 71.33\%&--&   --\\
\hline
Kim et al.~\cite{kim2016fusing}&\textbf{73.73\%}&--&   --\\
\hline
Bishay et al.~\cite{bishay2019schinet}  & -- & \textbf{73.1\%} &   --    \\
\hline
Lian et al.~\cite{lianexpression}  & -- & 71.9 \% &   --    \\
\hline
Acharya et al.~\cite{acharya2018covariance} &-- & --  & \textbf{87\%}  \\
\hline
S Li et al.~\cite{li2018reliable} &-- & --  &74.2\%  \\
\hline
Z.Liu et al.~\cite{liu2017boosting} &-- & --  &73.19\%  \\
\hline
\multicolumn{4}{l}{$^{\mathrm{a}}$KDL:  Fully connected kernel.}
\end{tabular}
}
\label{tab:SOTA}
\end{center}
\end{table}

\subsection{Comparison with the State-of-the-Art}

In this section, we compare the performance of the KDL with CNN to several state-of-the-art FER methods. The obtained results are reported in Table~\ref{tab:SOTA}. As can be seen, KDL with CNN outperforms the state-of-the-art methods on the ExpW dataset. The best accuracy rate is 76.64\% and has been reached using the third-order polynomial KDL. Second-order polynomial KDL gives, for his turn, 76.13\%. Whereas the state-of-the-art methods~\cite{bishay2019schinet} reached \textbf{73.1\%}.

On RAF-DB dataset, the accuracy of our models is also superior to state-of-the-art methods. The best accuracy rate is 88.02\% and has been reached using the third-order polynomial KDL. Second-order polynomial KDL gives, for his turn, 87.64\%. Whereas the state-of-the-art methods~\cite{acharya2018covariance} gives \textbf{87\%}.

For FER2013, even thought using the KDL improves considerably the models accuracy, the obtained results are still under the state-of-the-art results. The best accuracy rate for this dataset, namely 71.28\%, was reached using third-order KDL. Despite the improvement, the result obtained is 2.5\% less than the state-of-the-art method~\cite{kim2016fusing} (\textbf{73.73\%}) but remains competitive.

\section{Conclusion}
\label{sec:Conclusion}

In this paper, we designed Kernelized Dense Layer for CNN model that aims to enhance the discriminative power of the overall model. It consists of applying higher order kernel method than the standard FC layer. Experimental results on ExpW, RAF-DB and FER2013 datasets demonstrate the efficiency of the proposed KDL compared to standard FC layer in terms of convergence, speed and overall accuracy. The proposed FER method outperforms  most of the state-of-the-art methods and remains competitive. The performance of our model is essentially due to its capability of capturing high order information that are crucial for fine-grained classification tasks such as the FER.

As future work, other kernel functions will be considered, compared and combined. Different configurations will be studied, especially  the impact of the number of KDL used in the network.


\bibliographystyle{IEEEbib}
\bibliography{main}

\end{document}